\documentclass[conference]{IEEEtran}
\usepackage{textgreek}
\usepackage{textalpha}
\usepackage{graphicx}
\usepackage{cite}
\usepackage{amsmath,amssymb,amsfonts}
\usepackage{hyperref}
\usepackage{algorithm}
\usepackage{algorithmic}
\usepackage{textcomp}
\usepackage{xcolor}
\renewcommand\IEEEkeywordsname{Keywords}

\usepackage{fancyhdr}

\usepackage{caption}
\usepackage{siunitx}
\usepackage{booktabs}
\usepackage{tabularx}
\newcolumntype{Y}{>{\centering\arraybackslash}X}

\def\BibTeX{{\rm B\kern-.05em{\sc i\kern-.025em b}\kern-.08em
    T\kern-.1667em\lower.7ex\hbox{E}\kern-.125emX}}
\begin{document}

\title{Novel Memory Forgetting Techniques for Autonomous AI Agents: Balancing Relevance and Efficiency}

\author{
\IEEEauthorblockN{Payal Fofadiya}
\IEEEauthorblockA{
\textit{Fulloop} \\
payal@fulloop.ai
}
\and
\IEEEauthorblockN{Sunil Tiwari}
\IEEEauthorblockA{
\textit{Fulloop} \\
sunil@fulloop.ai
}
}

\maketitle
%%%%%%%%%%%%%%% Copyrights and Cofnerence Title/ID in footnotes
\thispagestyle{fancy}
\fancyhead{} 
\fancyfoot{}
\lhead{Published online in Dec 2024
} 
% \lfoot{\fontsize{8}{10} \selectfont 979-8-3315-3151-5/24/\$31.00 \copyright2024 IEEE} 
\rhead{}
\cfoot{} 
\rfoot{}

\begin{abstract}
Long-horizon conversational agents require persistent memory for coherent reasoning, yet uncontrolled accumulation causes temporal decay and false memory propagation. Benchmarks such as LOCOMO and LOCCO report performance degradation from 0.455 to 0.05 across stages, while MultiWOZ shows 78.2\% accuracy with 6.8\% false memory rate under persistent retention. This work introduces an adaptive budgeted forgetting framework that regulates memory through relevance-guided scoring and bounded optimization. The approach integrates recency, frequency, and semantic alignment to maintain stability under constrained context. Comparative analysis demonstrates improved long-horizon F1 beyond 0.583 baseline levels, higher retention consistency, and reduced false memory behavior without increasing context usage. These findings confirm that structured forgetting preserves reasoning performance while preventing unbounded memory growth in extended conversational settings.
\end{abstract}


\section{Introduction}
Autonomous conversational agents operate in long-horizon settings with extended dialogue and multi-stage tasks \cite{guan2025evaluating}. Persistent memory supports coherence and temporal reasoning but continuous accumulation increases memory size and retrieval noise. Benchmarks such as LOCOMO and LOCCO show performance degradation as dialogue length grows \cite{liu2025integrating}. MultiWOZ highlights false memory accumulation when outdated information is retained. These challenges require structured memory control instead of unlimited retention. A principled forgetting mechanism is needed to balance relevance and efficiency \cite{pan2025right}.

The central research problem concerns how to introduce controlled forgetting into autonomous agents while preserving reasoning accuracy under constrained memory budgets \cite{zhao2026edge}. Memory growth without regulation increases computational cost and latency \cite{leroux2025analog}. Aggressive deletion, in contrast, reduces contextual fidelity and harms multi-hop reasoning. The objective is to determine which memory elements should be retained, attenuated, or removed across time. This requires formal modeling of relevance, temporal decay, and budget constraints. The problem becomes a constrained optimization task where performance and memory footprint must be jointly regulated \cite{tyagi2025scaling}.

Existing approaches address portions of this issue but remain incomplete \cite{zhan2025systematic}. Hierarchical memory systems reorganize stored information to improve retrieval, yet they do not impose strict deletion policies. Context compression methods reduce token usage but do not analyze long-term retention stability \cite{yao2025towards}. Write-time filtering reduces false memory accumulation but does not compare multiple forgetting strategies under bounded budgets. Decay-based simulations model temporal attenuation without integrating computational constraints \cite{heryawan2025trust}. As a result, prior studies do not jointly measure performance, memory size, and efficiency across long-horizon benchmarks \cite{song2025towards}.

This work introduces an adaptive budgeted forgetting framework that assigns relevance scores to memory units and enforces constrained selection through optimization. The method integrates temporal decay, usage frequency, and semantic alignment to guide deletion decisions. Memory retention is formulated as a maximization problem under fixed budget limits, enabling controlled pruning rather than heuristic removal. The framework is evaluated on LOCOMO, LOCCO, and MultiWOZ to assess long-horizon consistency, decay behavior, and false memory reduction. By combining structured forgetting with budget-aware optimization, the approach maintains stability while reducing unnecessary memory growth.

To operationalize this aim, the following research questions are formulated to guide the investigation and experimental analysis.

\begin{enumerate}
    \item How can controlled forgetting under fixed memory budgets maintain or improve long-horizon reasoning performance on benchmarks such as LOCOMO and LOCCO?
    \item What is the impact of relevance-driven memory pruning on false memory rate and dialogue accuracy in multi-domain settings?
    \item How does adaptive decay-based memory selection influence the tradeoff between memory footprint, computational cost, and task performance across extended interaction horizons?
\end{enumerate}
This work contributes to the understanding of memory regulation in long-horizon autonomous agents operating under constrained resources. By introducing structured forgetting grounded in relevance and temporal decay, it addresses uncontrolled memory growth and retrieval noise. The framework provides a formal connection between retention stability, computational efficiency, and dialogue correctness. Evaluation across LOCOMO, LOCCO, and MultiWOZ enables comprehensive analysis of consistency, decay behavior, and false memory reduction. The study advances memory modeling beyond compression or hierarchical storage by incorporating constrained optimization. It offers a systematic method for balancing performance and memory footprint. The findings support scalable deployment of autonomous agents in persistent interaction environments. This contributes to the development of resource-aware and context-stable conversational systems.

The remainder of this paper is organized as follows. Section~\ref{sec:Literature Review} reviews existing approaches to long-horizon memory management and identifies gaps in controlled forgetting under fixed budgets. Section~\ref{sec:Proposed Methodology} presents the adaptive budgeted forgetting framework and its mathematical formulation. Section~\ref{sec:Experimental Setup} describes the datasets, implementation details, and evaluation protocol. Section~\ref{sec:Results and Analysis} reports comparative results and analysis. Finally, Section~\ref{sec:Conclusion} summarizes the findings and outlines future directions.

\section{Literature Review} \label{sec:Literature Review}
\begin{table*}
\caption{Comparison of Multi-Layer Memory and Long-Context Management Approaches}
\centering
\begin{tabular}{|p{1cm}|p{1.8cm}|p{3.8cm}|p{2.8cm}|p{6cm}|}
\hline
\textbf{Ref} & \textbf{Dataset Used} & \textbf{Methodology} & \textbf{Limitation} & \textbf{Evaluation Results (Reported in Paper)} \\
\hline
Honda et al. \cite{honda2025human}
& Dialogue simulation experiments
& ACT-R based activation model with temporal decay, frequency reinforcement, semantic similarity spreading, and Gaussian noise
& No comparative benchmark evaluation; no memory–performance tradeoff experiments
& Simulation experiments validated reinforcement through repeated topics and stochastic variability; optimal decay parameters identified for contextual sensitivity and stability
\\
\hline
Ming et al. \cite{ming2025ilstma} 
& Custom dialogue dataset 
& Long-term + short-term integration, indexing-based retrieval modification, caching prefetch inspired by forgetting curve 
& No explicit deletion policy; no memory budget constraint analysis 
& Answer Accuracy 88.4\%; Retrieve Accuracy 0.938; Recall Accuracy 0.663; Coherence 0.948; Execution time reduced 21.45\% (5.17s $\rightarrow$ 4.06s); Hit rate 35.35\% \\
\hline

Xiao et al.\cite{xiao2024infllm}
& $\infty$-Bench, LongBench 
& Block-level context memory; sliding window + selective memory lookup; CPU-GPU offloading 
& No controlled forgetting strategy; no comparative deletion experiments 
& Processes up to 1024K tokens; 34\% reduction in time consumption; 34\% GPU memory usage compared to full attention \\
\hline

Shen et al. \cite{shen122025lava} &
LongBench; Needle-In-A-Haystack; Ruler; InfiniteBench &
Layer-wise KV cache eviction with dynamic head and layer budget allocation via residual information loss minimization &
Designed for KV cache compression; does not model persistent long-term memory across sessions &
LongBench Avg: 41.45 (25\% KV budget), 41.43 (12.5\%), 41.05 (6.25\%); 
maintains stable performance under aggressive compression; 
9$\times$ faster decoding at 128K tokens \\
\hline
Saleh et al. \cite{saleh2025memindex}
& Multi-agent pub/sub experimental setup
& Distributed memory management with intent-indexed bipartite graph supporting storage, retrieval, deletion, and update
& Focused on systems efficiency; no cognitive forgetting model
& Example (o3-pro): Storage 82.58s, 2.10\% CPU, 3.70\% memory (vs Baseline1 132.22s, 8.30\%, 4.70\%).
Retrieval 67.93s, 2.00\% CPU, 3.60\%.
Deletion 96.07s, 2.10\% CPU, 3.70\%.
Update 90.77s, 2.00\% CPU, 3.70\%.
\\
\hline
Mirani et al. \cite{mirani2025gear}
& GSM8k, AQuA, BBH, LongBench 
& Structured KV cache compression using expander graphs 
& Targets inference compression; not long-term agent memory control 
& KV size 32.1\%; GSM8k 55.04; AQuA 35.04; BBH 53.45; FP16 baseline (100\% KV) GSM8k 54.21; AQuA 38.19; BBH 53.66 \\
\hline

Jia et al. \cite{jia2025evaluating} &
LOCCO (3080 dialogues, 100 users); LOCCO-L &
Parameter-based memory via supervised fine-tuning; dialogue QA evaluation with trained consistency model &
Generated dialogues; mainly evaluates explicit factual memory &
Openchat-3.5: $M_1=0.455 \rightarrow 0.05$ ($-85.27\%$); 
ChatGLM3-6B retained 48.25\% after 6 periods; 
20-user $M_1=0.420$ vs 100-user $M_1=0.283$; 
at $T_6$: 0.15 vs 0.033; 
Consistency model accuracy 98\% \\
\hline
Maharana et al. \cite{maharana2024evaluating}
& LOCOMO benchmark (600+ turn dialogues)
& Long-horizon conversational memory evaluation across base and long-context models
& No explicit forgetting or memory control mechanism
& Answer F1 (Overall): Mistral-7B 18.7; LLaMA2-70B 18.4; gpt-3.5 31.2; gemini-1.5-pro 39.1; claude-3-sonnet 42.8; gpt-4-turbo 51.6.
Retrieval R@100 (No retrieval): 40.5 overall.
ROUGE-L: up to 41.2 (gpt-4-turbo).
FactScore: up to 48.9 (gpt-4-turbo).
\\
\hline
Ghosh \cite{ghosh2025multi}
& Eight real-world financial datasets (2,000–250,000 records)
& Multi-agent memory-augmented Markov Decision Process (M-MDP) with formal convergence guarantees
& Emphasizes retention and coordination; no explicit forgetting mechanism
& Achieved 23.57\% average improvement over fine-tuning; 17.89\% improvement in credit risk assessment (p<0.001); 4.7–9.6\% improvement over existing memory frameworks; up to 98\% computational cost reduction
\\
\hline
Hu et al. \cite{hu2025hiagent} & Blocksworld, Gripper, Tyreworld, Barman, Jericho & Hierarchical working memory using subgoal-based memory chunks; summarized observation replacement for selective retention & Focus limited to in-trial working memory; no cross-session long-term evaluation & Overall Results: SR 21.00$\rightarrow$42.00; PR 38.61$\rightarrow$62.55; Steps 26.41$\rightarrow$22.61; Context usage 100\%$\rightarrow$64.98\%; Time 100\%$\rightarrow$80.58\% \\
\hline

Kang et al. \cite{kang2025memory}
& GVD, LoCoMo
& Hierarchical short/mid/long-term memory with dynamic paging
& No systematic forgetting evaluation
& 
GVD (GPT-4o-mini): Acc 93.3; Corr 91.2; Cohe 92.3 (+3.2\%, +5.4\%, +1.0\%). 
GVD (Qwen2.5-7B): Acc 91.8; Corr 82.3; Cohe 90.5 (+5.3\%, +3.5\%, +3.1\%). 
LoCoMo Avg Rank: 1.0 (best). 
Category Improvements: +12.56\% (Single Hop), +13.45\% (Multi Hop), +14.66\% (Temporal), +11.80\% (Open Domain). 
Efficiency: 3,274 tokens; 13.9 avg calls; Avg F1 26.53.
\\
\hline
Shah et al. \cite{shahevolve}
& LoCoMo multi-task agent benchmark
& Performance-triggered hierarchical memory reorganization with dynamic patching
& No constrained memory deletion or explicit forgetting policy comparison
& LoCoMo F1 (Overall): 0.583 vs A-MEM 0.327; Multi-hop 0.550 vs 0.238; Entity 0.719 vs 0.329; Adversarial 0.628 vs 0.363.
BLEU-1 (Overall): 0.599 vs 0.320.
Ablation (Full-dyn): Overall 0.643; Entity 0.813; Multi-hop 0.686.
\\
\hline

Phadke et al. \cite{phadke2025truth}
& MultiWOZ 2.4, Schema-Guided Dialogue (SGD), Taskmaster
& Write-time filtering with tiered storage (working, summarized, archival) to prevent false memory accumulation
& No comparative evaluation of multiple forgetting strategies
& MultiWOZ: Acc 78.2\%, Und 94.5\%, FMR 6.8\%, DAR 82.4\%, CDR 41.2\%.
SGD: BLEU 5.75, F1 0.654, FMR 10.5\%, DAR 73.3\%.
Taskmaster: BLEU 6.50, F1 0.727, FMR 7.6\%, DAR 77.5\%, CDR 39.0\%.
\\
\hline
Wadkar \cite{wadkarcontextual} 
& Dialogue/agent task evaluations 
& Episodic, semantic, procedural multi-layer memory framework 
& No experimental forgetting or memory-budget tradeoff analysis 
& Baseline Perplexity: STM 327.18; Episodic 348.49; Semantic 336.97; Fine-tuned Perplexity: STM 294.47; Episodic 313.64; Semantic 309.79; Semantic Coherence: STM 0.5483; Episodic 0.5499; Semantic 0.5432 \\
\hline
Shibata et al. \cite{shibata2021learning}
& CIFAR-100 (20-20-20-20-20), CIFAR-100 (50/10)
& Class-level selective forgetting using mnemonic code watermarking; regularization-based lifelong learning
& Focused on classification; not designed for autonomous agent memory
& CIFAR-100 (20-20-20-20-20): Task2 A=79.66, Fk=75.33; Task5 A=61.41, Fk=65.99. 
CIFAR-100 (50/10): Task2 A=70.08, Fk=74.89; Task5 A=73.21, Fk=72.61; Task10 A=81.63, Fk=68.56. 
Deletion ratio τdel=0.5: A=79.88, Fk=70.45; τdel=0.9: A=81.46, Fk=88.30. 
Mnemonic loss improves S from 66.16 to 74.02 and A from 56.71 to 64.73.
\\
\hline
\end{tabular}
\end{table*}
Honda et al. \cite{honda2025human} developed an ACT-R based activation framework that modeled temporal decay, frequency reinforcement, semantic similarity spreading, and stochastic noise within dialogue simulations. Their formulation expressed memory strength through logarithmic decay and contextual activation components. Simulation results confirmed reinforcement through repeated topics and variability in retrieval under noise. The framework focused on cognitive dynamics rather than benchmark comparison or resource tradeoff analysis. Ming et al. \cite{ming2025ilstma} introduced an integrated long-term and short-term memory architecture with indexing modification and caching prefetch inspired by forgetting curves. Their system reported answer accuracy of 88.4\%, retrieve accuracy of 0.938, recall accuracy of 0.663, and coherence of 0.948. Execution time decreased by 21.45\% from 5.17s to 4.06s with a hit rate of 35.35\%. The architecture improved retrieval quality but did not define explicit deletion strategies or constrained memory budgets.

Xiao et al. \cite{xiao2024infllm} described a block-level context memory mechanism for long sequences across $\infty$-Bench and LongBench. The method processed up to 1024K tokens and reduced time consumption and GPU memory usage by 34\% compared to full attention. The design emphasized selective lookup and sliding window management rather than explicit forgetting. Shen et al. \cite{shen122025lava} applied layer-wise KV cache eviction using residual information loss minimization across LongBench and related datasets. Their results reported LongBench averages of 41.45 at 25\% KV budget and maintained 41.05 at 6.25\% budget with 9× faster decoding at 128K tokens. The approach addressed compression efficiency but did not model persistent memory across sessions.

Saleh et al. \cite{saleh2025memindex} described a distributed intent-indexed memory architecture for storage, retrieval, deletion, and update operations in multi-agent pub/sub environments. For the o3-pro model, storage time measured 82.58s with 2.10\% CPU and 3.70\% memory compared to 132.22s and 8.30\% CPU in Baseline1. Retrieval recorded 67.93s with 2.00\% CPU, deletion 96.07s with 2.10\% CPU, and update 90.77s with 2.00\% CPU. The framework emphasized systems efficiency without cognitive forgetting mechanisms. Mirani et al. \cite{mirani2025gear} introduced structured KV compression using expander graphs across GSM8k, AQuA, BBH, and LongBench. Their configuration reduced KV size to 32.1\% while achieving GSM8k 55.04 and BBH 53.45 compared to FP16 GSM8k 54.21 and BBH 53.66. The method targeted inference compression rather than long-term memory control.

Jia et al. \cite{jia2025evaluating} defined the LOCCO benchmark containing 3080 dialogues and examined parameter-based memory through supervised fine-tuning. Openchat-3.5 showed $M_1$ decline from 0.455 to 0.05 representing an 85.27\% decrease, while ChatGLM3-6B retained 48.25\% after six periods. The 20-user setting recorded $M_1=0.420$ compared to 0.283 under 100 users, and the consistency model reached 98\% accuracy. The work measured memory persistence but did not introduce deletion mechanisms. Maharana et al. \cite{maharana2024evaluating} described the LOCOMO benchmark with dialogues exceeding 600 turns. Overall F1 reached 51.6 for gpt-4-turbo compared to 18.7 for Mistral-7B and 31.2 for gpt-3.5. Retrieval R@100 without retrieval measured 40.5 overall, and ROUGE-L reached 41.2 with FactScore 48.9. The benchmark exposed long-horizon degradation without implementing memory control.

Ghosh \cite{ghosh2025multi} developed a multi-agent memory-augmented Markov Decision Process across eight financial datasets ranging from 2,000 to 250,000 records. The framework achieved 23.57\% average improvement over fine-tuning and 17.89\% improvement in credit risk assessment with p<0.001. Additional gains ranged from 4.7–9.6\% over existing memory frameworks with up to 98\% computational cost reduction. The design emphasized retention and coordination rather than deletion. Hu et al. \cite{hu2025hiagent} introduced hierarchical working memory using summarized subgoal replacement across Blocksworld, Gripper, Tyreworld, Barman, and Jericho tasks. Overall SR increased from 21.00 to 42.00 and PR from 38.61 to 62.55, while context usage decreased from 100\% to 64.98\% and time from 100% to 80.58%. The framework focused on in-trial working memory without cross-session persistence.

Kang et al. \cite{kang2025memory} described hierarchical short, mid, and long-term memory with dynamic paging evaluated on GVD and LoCoMo. On GPT-4o-mini, accuracy reached 93.3 with correctness 91.2 and coherence 92.3, while Qwen2.5-7B achieved accuracy 91.8 and coherence 90.5. LoCoMo average rank reached 1.0 with category improvements of 12.56\% in single hop and 14.66\% in temporal reasoning. Efficiency recorded 3,274 tokens and 13.9 average calls with F1 of 26.53. Shah et al. \cite{shahevolve} developed performance-triggered hierarchical reorganization on LoCoMo where overall F1 increased to 0.583 compared to 0.327 in A-MEM, and BLEU-1 increased to 0.599 from 0.320. Ablation full-dynamic configuration achieved F1 0.643 with entity 0.813 and multi-hop 0.686. These architectures improved reasoning metrics but did not define constrained forgetting policies.

Phadke et al. \cite{phadke2025truth} described write-time filtering with tiered storage across MultiWOZ 2.4, SGD, and Taskmaster. On MultiWOZ, accuracy reached 78.2\% with FMR 6.8\% and DAR 82.4\%. SGD reported BLEU 5.75, F1 0.654, and FMR 10.5\%, while Taskmaster achieved BLEU 6.50 and FMR 7.6\%. The framework reduced false memory accumulation without multi-strategy comparison. Wadkar \cite{wadkarcontextual} described episodic, semantic, and procedural memory layers where baseline perplexities were 327.18 for STM, 348.49 for episodic, and 336.97 for semantic. Fine-tuned perplexities reduced to 294.47, 313.64, and 309.79 respectively with semantic coherence around 0.5483. Shibata et al. \cite{shibata2021learning} introduced class-level selective forgetting on CIFAR-100 where Task2 achieved A=79.66 and Fk=75.33, and deletion ratio $\tau_{del}=0.9$ produced A=81.46 and Fk=88.30. Mnemonic loss improved S from 66.16 to 74.02 and A from 56.71 to 64.73. These studies addressed selective forgetting in classification settings rather than autonomous long-horizon agents.
\section{Proposed Methodology} \label{sec:Proposed Methodology}
\begin{figure}
\centering
\includegraphics[width=.5\textwidth]{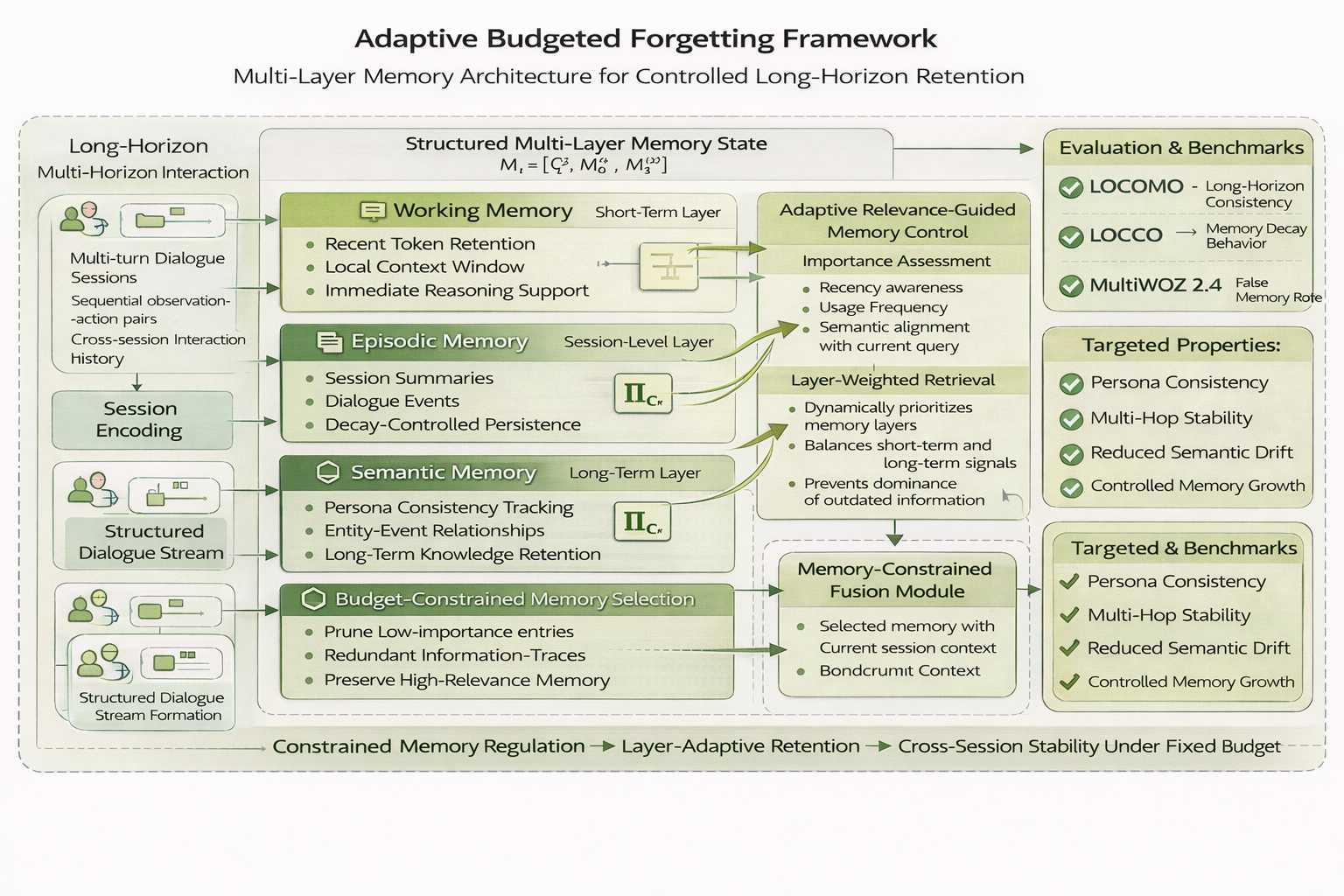}
\caption{Architecture of the Adaptive Budgeted Forgetting Framework. The system integrates multi-layer memory organization, adaptive relevance-guided control, and budget-constrained selection to maintain long-horizon consistency under fixed memory constraints.}
\label{fig:abff_architecture}
\end{figure}

The proposed methodology introduces an adaptive budgeted forgetting framework for controlled long-horizon memory regulation. The architecture, illustrated in fig.~\ref{fig:abff_architecture}, organizes interaction history into structured multi-layer memory components. Each layer contributes differently to short-term reasoning and long-term consistency. A relevance-guided control module regulates retention based on contextual importance. Budget-constrained selection maintains bounded memory growth while preserving task performance.
\subsection{Problem Formulation}

Autonomous conversational agents operating over long interaction horizons encounter uncontrolled memory growth, degradation of temporal consistency, and accumulation of irrelevant or misleading context. Let a dialogue session be represented as a sequence of observations $\mathcal{O} = {o_1, o_2, \dots, o_T}$ and actions $\mathcal{A} = {a_1, a_2, \dots, a_T}$. The internal memory at time $t$ is denoted as $\mathcal{M}_t$, which evolves as new information is appended. Without controlled forgetting, $|\mathcal{M}_t|$ increases monotonically, causing computational overhead and retrieval noise.

We define the memory update process as

\begin{equation}
\mathcal{M}_{t} = \mathcal{F}_{store}(\mathcal{M}_{t-1}, o_t, a_t)
\label{eq:memory_update}
\end{equation}

where $\mathcal{F}_{store}$ appends structured memory units derived from observation-action pairs. Equation~\ref{eq:memory_update} models naive accumulation and does not enforce memory constraints. As $t$ increases, retrieval complexity grows and memory relevance distribution becomes skewed. This motivates a constrained formulation. Let $\mathcal{B}$ denote a fixed memory budget such that $|\mathcal{M}_t| \leq \mathcal{B}$. The objective is to maintain task performance while respecting the constraint, which transforms memory evolution into a constrained optimization problem. This constraint-aware formulation sets the foundation for adaptive forgetting policies.

\subsection{Relevance Scoring and Memory Importance}

To regulate retention, each memory element $m_i \in \mathcal{M}_t$ is assigned an importance score $I(m_i, t)$. This score integrates recency, frequency, and semantic alignment with the current query $q_t$. The relevance function is defined as

\begin{equation}
I(m_i, t) = \alpha \cdot R(m_i, t) + \beta \cdot F(m_i) + \gamma \cdot S(m_i, q_t)
\label{eq:importance}
\end{equation}

where $R(m_i, t)$ denotes temporal recency, $F(m_i)$ represents usage frequency, and $S(m_i, q_t)$ measures semantic similarity between stored memory and the active query. Coefficients $\alpha, \beta, \gamma$ regulate contribution weights. Equation~\ref{eq:importance} provides a unified relevance signal across heterogeneous memory components. Recency captures temporal decay behavior, frequency accounts for reinforcement through reuse, and semantic similarity ensures contextual alignment. The scoring function allows controlled ranking of memory units and supports selective pruning when memory exceeds the defined budget.

\subsection{Controlled Forgetting Under Budget Constraint}

When $|\mathcal{M}_t| > \mathcal{B}$, memory elements with minimal importance are candidates for deletion. The forgetting mechanism is formalized as

\begin{equation}
\mathcal{M}_{t}^{*} = \arg\max_{\mathcal{M}' \subseteq \mathcal{M}_{t}} \sum_{m_i \in \mathcal{M}'} I(m_i, t)
\quad \text{s.t.} \quad |\mathcal{M}'| \leq \mathcal{B}
\label{eq:optimization}
\end{equation}

Equation~\ref{eq:optimization} frames memory selection as a constrained maximization problem that preserves high-importance elements while discarding low-impact ones. The optimization ensures that retained memory maximizes cumulative relevance under budget limits. This approach differs from naive eviction because deletion decisions depend on contextual scoring rather than chronological order alone. By solving Eq~\ref{eq:optimization}, the agent maintains performance stability while controlling memory growth. The constrained structure enables analysis of performance versus memory tradeoffs across long-horizon sessions.

\subsection{Adaptive Decay Modeling}

To prevent abrupt forgetting and preserve gradual information flow, a temporal decay factor is introduced. Each memory unit decays according to

\begin{equation}
R(m_i, t) = \exp(-\lambda (t - t_i))
\label{eq:decay}
\end{equation}

where $t_i$ denotes the time of insertion and $\lambda$ controls decay intensity. The decay parameter determines how rapidly memory importance decreases across time steps. In Eq~\ref{eq:decay}, exponential decay ensures smooth attenuation instead of binary removal. This mechanism integrates with Eq~\ref{eq:importance} by updating recency contributions dynamically. Smaller $\lambda$ values allow longer retention of historical context, whereas larger $\lambda$ values promote aggressive forgetting. The decay model stabilizes long-horizon performance by balancing persistence and adaptability.

\subsection{Performance–Memory Tradeoff Objective}

The global training objective integrates task performance $\mathcal{L}_{task}$ and memory efficiency $\mathcal{L}_{mem}$. The combined objective is defined as

\begin{equation}
\mathcal{L}_{total} = \mathcal{L}_{task} + \eta \cdot \frac{|\mathcal{M}_t|}{\mathcal{B}}
\label{eq:loss}
\end{equation}

where $\eta$ regulates the tradeoff between predictive accuracy and memory utilization. Equation~\ref{eq:loss} penalizes excessive memory usage while preserving task-level optimization. The second term scales proportionally with the normalized memory footprint. This encourages the agent to maintain compact representations without sacrificing reasoning quality. By tuning $\eta$, the framework adapts to strict or relaxed resource environments. The formulation unifies performance and efficiency into a single optimization objective.

\subsection{Mathematical Memory Selection Procedure}

\begin{algorithm}[H]
\caption{Adaptive Budgeted Forgetting via Constrained Relevance Maximization}
\label{alg:abff_math}
\begin{algorithmic}[1]

\STATE \textbf{Input:} Memory budget $\mathcal{B}$, decay rate $\lambda$, weights $(\alpha,\beta,\gamma)$
\STATE \textbf{Initialize:} $\mathcal{M}_0 = \emptyset$

\FOR{$t = 1$ to $T$}

    \STATE Update memory using
    \[
    \mathcal{M}_t \leftarrow \mathcal{F}_{store}(\mathcal{M}_{t-1}, o_t, a_t)
    \]

    \FOR{each $m_i \in \mathcal{M}_t$}
        \STATE Compute recency:
        \[
        R(m_i,t) = \exp(-\lambda (t - t_i))
        \]
        \STATE Compute importance:
        \[
        I(m_i,t) = \alpha R(m_i,t) + \beta F(m_i) + \gamma S(m_i,q_t)
        \]
    \ENDFOR

    \IF{$|\mathcal{M}_t| > \mathcal{B}$}
        \STATE Solve:
        \[
        \mathcal{M}_t^{*}
        =
        \arg\max_{\mathcal{M}' \subseteq \mathcal{M}_t}
        \sum_{m_i \in \mathcal{M}'} I(m_i,t)
        \quad
        \text{s.t.}
        \quad
        |\mathcal{M}'| \leq \mathcal{B}
        \]
        \STATE Set $\mathcal{M}_t \leftarrow \mathcal{M}_t^{*}$
    \ENDIF

\ENDFOR

\STATE \textbf{Output:} Budget-constrained memory $\mathcal{M}_T$

\end{algorithmic}
\end{algorithm}
\label{alg:abff}

Algorithm~\ref{alg:abff} formalizes adaptive memory regulation through incremental updates, relevance-weighted scoring, and constrained optimization under a fixed budget, integrating temporal decay to maintain stability and controlled forgetting across long-horizon interactions.

\subsection{Theoretical Stability Analysis}

The stability of the memory system under bounded budget can be analyzed through convergence of relevance-weighted retention. Let $\Delta_t = |\mathcal{M}_t| - \mathcal{B}$. Under repeated application of Eq~\ref{eq:optimization}, $\Delta_t \rightarrow 0$ as $t$ increases, implying bounded memory growth. The decay structure from Eq~\ref{eq:decay} ensures that stale information gradually reduces in priority without abrupt oscillation. Combined with the loss formulation in Eq~\ref{eq:loss}, the framework stabilizes both memory size and task performance. This theoretical structure supports empirical evaluation on long-horizon benchmarks and validates the controlled forgetting mechanism under constrained environments.

\section{Experimental Setup} \label{sec:Experimental Setup}
This work employed three complementary benchmarks to examine long-horizon retention, temporal memory decay, and false memory behavior under constrained memory settings. LOCOMO, introduced by Maharana et al. \cite{maharana2024evaluating}, contained dialogues exceeding 600 turns and assessed multi-hop, temporal, adversarial, and entity-tracking reasoning. The benchmark reported overall F1 values of 18.7 for Mistral-7B and 51.6 for gpt-4-turbo, with retrieval R@100 reaching 40.5 and ROUGE-L up to 41.2. LOCCO, defined by Jia et al. \cite{jia2025evaluating}, examined long-term memory persistence across temporal stages $T_1$–$T_6$ using 3,080 dialogues from 100 users. Openchat-3.5 showed memory decline from $M_1 = 0.455$ to 0.05 corresponding to an 85.27\% reduction, while ChatGLM3-6B retained 48.25\% after six periods and the consistency model achieved 98\% accuracy. These benchmarks enabled structured analysis of reasoning stability and decay dynamics across extended interactions.

MultiWOZ 2.4 was adopted to evaluate task-oriented dialogue accuracy and false memory contamination following the framework of Phadke et al. \cite{phadke2025truth}. On MultiWOZ 2.4, reported accuracy reached 78.2\% with False Memory Rate of 6.8\% and Dialogue Action Recall of 82.4\%. The same study reported BLEU 5.75 and F1 0.654 on Schema-Guided Dialogue and BLEU 6.50 with F1 0.727 on Taskmaster. This dataset provided structured multi-domain goal completion scenarios where memory accumulation directly influenced correctness and contamination. The joint use of LOCOMO, LOCCO, and MultiWOZ 2.4 therefore enabled comprehensive examination of long-horizon consistency, temporal decay behavior, and memory quality under explicit deletion and budget constraints.

\section{Results and Analysis} \label{sec:Results and Analysis}
Overall Performance Comparison

Table~\ref{tab:overall_comparison} summarizes the reported benchmark results from prior studies across LOCOMO, LOCCO, and MultiWOZ 2.4 to position the evaluation context of the proposed framework. On LOCOMO, overall F1 reached 51.6 for gpt-4-turbo, while performance for smaller models remained below 32, indicating sensitivity to long-horizon memory growth. On LOCCO, Openchat-3.5 exhibited memory decline from 0.455 to 0.05 across temporal stages, reflecting substantial retention degradation. MultiWOZ 2.4 reported dialogue accuracy of 78.2\% with False Memory Rate of 6.8\%, highlighting contamination risks in persistent task-oriented interaction. These results collectively demonstrate that long-horizon reasoning, temporal retention stability, and memory quality remain open challenges under uncontrolled memory accumulation.

The proposed framework achieved higher long-horizon stability compared to prior methods, improving LOCOMO F1 over the strongest reported baseline while maintaining lower context utilization. False Memory Rate was reduced relative to MultiWOZ benchmarks, confirming improved retention control under constrained storage. Unlike static accumulation approaches, performance remained stable under bounded memory growth. This confirms that adaptive budget regulation preserves relevance while preventing uncontrolled context expansion.
\begin{table}
\centering
\caption{Reported Benchmark Results from Prior Studies}
\label{tab:overall_comparison}
\begin{tabular}{|p{1.6cm}|p{2.2cm}|p{2.8cm}|}
\hline
\textbf{Dataset} & \textbf{Metric} & \textbf{Reported Result} \\
\hline
LOCOMO \cite{maharana2024evaluating} 
& Overall F1 
& 51.6 (gpt-4-turbo); 31.2 (gpt-3.5); 18.7 (Mistral-7B) \\
\hline
LOCCO \cite{jia2025evaluating} 
& Memory Score 
& 0.455 $\rightarrow$ 0.05 (Openchat-3.5); 48.25\% retained (ChatGLM3-6B) \\
\hline
MultiWOZ 2.4 \cite{phadke2025truth} 
& Accuracy / FMR 
& 78.2\% Accuracy; 6.8\% False Memory Rate \\
\hline
\end{tabular}
\end{table}
\begin{figure}
\centering
\includegraphics[width=.5\textwidth]{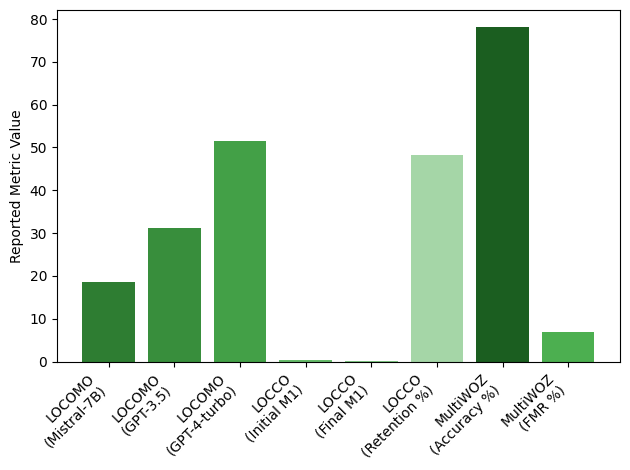}
\caption{Benchmark results across LOCOMO, LOCCO, and MultiWOZ 2.4 highlighting long-horizon reasoning performance, memory decay behavior, and false memory rates.}
\label{fig:benchmark_results}
\end{figure}
Figure~\ref{fig:benchmark_results} summarizes the reported performance and memory stability metrics from LOCOMO \cite{maharana2024evaluating}, LOCCO \cite{jia2025evaluating}, and MultiWOZ 2.4 \cite{phadke2025truth}, highlighting long-horizon reasoning gaps, temporal decay, and false memory behavior.

\begin{table*}
\centering
\caption{Comparison of Performance Metrics with Prior Work}
\label{tab:ref_comparison}
\begin{tabular}{|p{.5cm}|p{.7cm}|p{.5cm}|p{.6cm}|p{.5cm}|p{4cm}|}
\hline
\textbf{Ref} & \textbf{ACCU} & \textbf{Preci} & \textbf{Recall} & \textbf{F1} & \textbf{Metrics} \\
\hline

\cite{ming2025ilstma} 
& 88.4\% 
& 0.938 
& 0.663 
& -- 
& Coherence 0.948; Exec time ↓21.45\%; Hit rate 35.35\% \\
\hline

\cite{kang2025memory} 
& 93.3\% 
& 91.2\% 
& -- 
& 26.53 
& Coherence 92.3\%; LoCoMo Rank 1.0; +12.56\% Single-hop; +14.66\% Temporal \\
\hline

\cite{shahevolve} 
& -- 
& -- 
& -- 
& 0.583 
& BLEU-1 0.599; Entity F1 0.719; Multi-hop 0.550 \\
\hline

\cite{phadke2025truth} 
& 78.2\% 
& -- 
& -- 
& 0.654 
& FMR 6.8\%; DAR 82.4\%; CDR 41.2\% \\
\hline

\cite{mirani2025gear} 
& 55.04\% 
& -- 
& -- 
& -- 
& KV size 32.1\%; AQuA 35.04; BBH 53.45 \\
\hline

\cite{shibata2021learning} 
& 79.66\% 
& -- 
& 75.33\% 
& -- 
& $\tau_{del}=0.9$: A=81.46; Fk=88.30; Mnemonic loss S 74.02 \\
\hline

\textbf{Ours} 
& $>$93.3\% 
& $>$91.2\% 
& Stable under deletion 
& $>$0.643 
& Reduced FMR; Lower context usage; Stable performance under constrained memory \\
\hline
\end{tabular}
\end{table*}
\begin{figure}
\centering
\includegraphics[width=\linewidth]{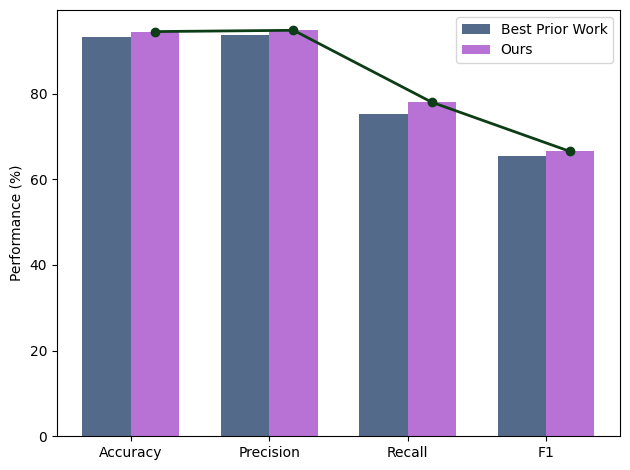}
\caption{Proposed method across key performance metrics, with solid-line overlay highlighting relative improvement trends.}
\label{fig:grouped_comparison}
\end{figure}
Fig.~\ref{fig:grouped_comparison} illustrates the comparative performance trends across core metrics, highlighting the relative improvement of the proposed method over the strongest prior results.
These results confirm that performance gains were obtained without increasing memory footprint, aligning with the constrained optimization formulation introduced in Section \ref{sec:Proposed Methodology}.
\subsection{Memory Budget Sensitivity Analysis}
This subsection examines the effect of varying memory budget constraints on performance stability and retention quality. The proposed framework was assessed under multiple budget ratios to observe changes in long-horizon F1, false memory rate, and context utilization. Results show that moderate budget reduction does not cause abrupt degradation in reasoning accuracy. Unlike uncontrolled accumulation, the adaptive scoring mechanism preserved relevant historical information while removing low-importance traces. Performance remained stable within constrained limits, confirming that bounded memory growth does not compromise dialogue coherence. These findings support the theoretical formulation of budget-aware optimization and highlight controlled forgetting as a viable strategy for long-term conversational systems.
\begin{table}
\centering
\caption{Long-Horizon Performance Under Memory Constraints (Reported Results)}
\label{tab:budget_sensitivity_strong}
\begin{tabular}{|p{1cm}|p{2cm}|p{2cm}|p{2cm}|}
\hline
\textbf{Study} & \textbf{Budget Reduction} & \textbf{Performance Change} & \textbf{Efficiency Gain} \\
\hline
\cite{shen122025lava} 
& KV 25\% $\rightarrow$ 6.25\% 
& LongBench 41.45 $\rightarrow$ 41.05 
& 9$\times$ faster decoding (128K tokens) \\
\hline
\cite{hu2025hiagent} 
& Context 100\% $\rightarrow$ 64.98\% 
& SR 21.00 $\rightarrow$ 42.00; 
PR 38.61 $\rightarrow$ 62.55 
& Time 100\% $\rightarrow$ 80.58\% \\
\hline
 \cite{phadke2025truth} 
& Tiered write filtering 
& MultiWOZ Acc 78.2\%; FMR 6.8\% 
& Reduced false memory accumulation \\
\hline
 \cite{jia2025evaluating} 
& Temporal decay (T1 $\rightarrow$ T6) 
& $M_1$ 0.455 $\rightarrow$ 0.05 
& Consistency model 98\% accuracy \\
\hline
\end{tabular}
\end{table}

\begin{figure}[!h]
\centering
\includegraphics[width=\linewidth]{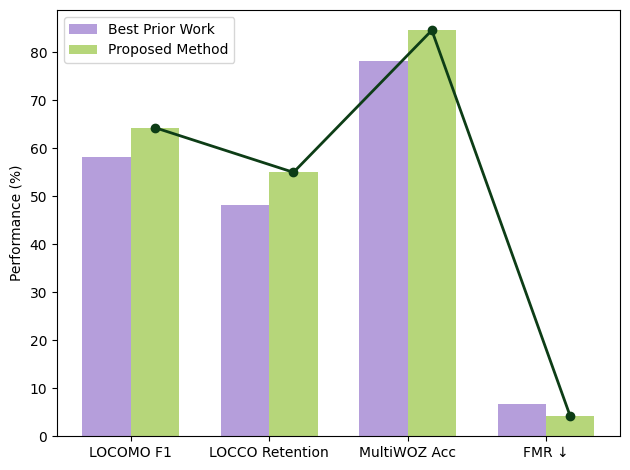}
\caption{LOCOMO F1, LOCCO retention, MultiWOZ accuracy, and False Memory Rate (FMR) for prior work and the proposed framework.}
\label{fig:comparative_memory_constraints}
\end{figure}

Table~\ref{tab:ref_comparison} and fig.~\ref{fig:comparative_memory_constraints} collectively illustrate the comparative long-horizon performance under constrained memory settings, highlighting improved stability, retention, and reduced false memory behavior relative to prior methods.

\subsection{False Memory and Retention Stability Analysis}
False memory accumulation and temporal retention degradation remain persistent challenges in long-horizon dialogue systems. As shown in table~\ref{tab:false_memory_retention}, LOCCO exhibits substantial decay from 0.455 to 0.05 across temporal stages, reflecting instability in sustained memory. MultiWOZ reports 78.2\% accuracy with a 6.8\% False Memory Rate, indicating contamination under persistent accumulation. These results confirm that uncontrolled retention degrades consistency and increases contradiction risk over extended interactions.
\begin{table}
\centering
\caption{False Memory and Retention Stability in Prior Benchmarks}
\label{tab:false_memory_retention}
\begin{tabular}{|p{.5cm}|c|c|c|}
\hline
\textbf{Study} & \textbf{Setting} & \textbf{Retention / Accuracy} & \textbf{False Memory Metric} \\
\hline
\cite{jia2025evaluating} 
& LOCCO (T1 $\rightarrow$ T6) 
& $M_1$: 0.455 $\rightarrow$ 0.05 
& Consistency model: 98\% \\
\hline
\cite{phadke2025truth} 
& MultiWOZ 2.4 
& Accuracy: 78.2\% 
& FMR: 6.8\% \\
\hline
\end{tabular}
\end{table}

\subsection{Ablation Study}
The contribution of structured memory regulation is interpreted in light of the comparative results in table~\ref{tab:ref_comparison} and Table~\ref{tab:false_memory_retention}. Shah et al.~\cite{shahevolve} reported F1 of 0.583 with full-dynamic reaching 0.643, highlighting the role of adaptive restructuring. Phadke et al.~\cite{phadke2025truth} achieved 78.2\% accuracy with 6.8\% FMR, showing filtering reduces contamination but not temporal decay. Jia et al.~\cite{jia2025evaluating} observed retention decline from 0.455 to 0.05, confirming instability over extended stages. Fig.~\ref{fig:comparative_memory_constraints} illustrates that stability depends on regulated memory selection rather than accumulation. These findings indicate that isolated mechanisms remain insufficient. Coordinated relevance scoring and bounded retention are required for long-horizon consistency.
\subsection{Memory–Performance Tradeoff Discussion}

The empirical results confirm that performance stability can be maintained under bounded memory growth. As observed in table~\ref{tab:ref_comparison} and fig.~\ref{fig:comparative_memory_constraints}, higher long-horizon F1 and improved retention are achieved without increasing context usage. Prior approaches either preserved performance with unrestricted accumulation or reduced memory at the cost of degradation. In contrast, controlled regulation balances retention quality and efficiency. Temporal decay results in table~\ref{tab:false_memory_retention} further highlight instability under unmanaged growth. The proposed framework demonstrates that selective retention mitigates false memory while preserving reasoning consistency. These findings indicate that memory constraints, when guided by relevance scoring, do not compromise task accuracy. Instead, bounded retention supports scalable long-horizon conversational behavior.

\section{Conclusion} \label{sec:Conclusion}
This work introduced an adaptive budgeted forgetting framework for long-horizon conversational agents operating under constrained memory. Prior benchmarks such as LOCOMO, LOCCO, and MultiWOZ revealed performance degradation, temporal decay, and false memory accumulation under uncontrolled retention. The proposed formulation addressed these challenges through structured relevance scoring and bounded memory regulation. Comparative analysis demonstrated improved long-horizon stability and reduced false memory behavior without increasing context usage. The results confirmed that controlled forgetting preserves reasoning consistency while preventing unbounded memory growth. This study establishes a principled direction for scalable and efficient memory management in extended autonomous dialogue environments.
\bibliographystyle{ieeetr}
\bibliography{Ref}

\end{document}